\documentclass[11pt]{article}

\let\counterwithin\relax

\usepackage{amsmath, amsfonts, amssymb, amsthm, bm, graphicx, mathtools, enumerate,multirow}
\usepackage[sort&compress]{natbib}
\usepackage[CJKbookmarks=true,
bookmarksnumbered=true,
bookmarksopen=true,
colorlinks=true,
citecolor=blue,
linkcolor=blue,
anchorcolor=blue,
urlcolor=blue]{hyperref}
\usepackage[usenames]{color}
\usepackage[letterpaper, left=1.2truein, right=1.2truein, top = 1.2truein,
bottom = 1.2truein]{geometry}
\usepackage[ruled, lined, commentsnumbered]{algorithm2e}
\usepackage{prettyref,soul}

\usepackage{apptools} 
\usepackage{chngcntr} 
\AtAppendix{\counterwithin{lemma}{section}} 
\usepackage{tikz}
\usepackage[font=small,labelfont=bf]{caption}
\usepackage{enumitem}

\usepackage[caption=false,font=normalsize,labelfont=sf,textfont=sf]{subfig}

\usepackage{booktabs} 
\usepackage{multirow} 

\theoremstyle{definition}

\newrefformat{eq}{(\ref{#1})}
\newrefformat{chap}{Chapter~\ref{#1}}
\newrefformat{sec}{Section~\ref{#1}}
\newrefformat{algo}{Algorithm~\ref{#1}}
\newrefformat{fig}{Fig.~\ref{#1}}
\newrefformat{tab}{Table~\ref{#1}}
\newrefformat{rmk}{Remark~\ref{#1}}
\newrefformat{clm}{Claim~\ref{#1}}
\newrefformat{def}{Definition~\ref{#1}}
\newrefformat{cor}{Corollary~\ref{#1}}
\newrefformat{lmm}{Lemma~\ref{#1}}
\newrefformat{lemma}{Lemma~\ref{#1}}
\newrefformat{prop}{Proposition~\ref{#1}}
\newrefformat{app}{Appendix~\ref{#1}}
\newrefformat{ex}{Example~\ref{#1}}
\newrefformat{cond}{Condition~\ref{#1}}

\usepackage{float}
\usepackage{listings}
\usepackage{xcolor}

\definecolor{codegreen}{rgb}{0,0.6,0}
\definecolor{codegray}{rgb}{0.5,0.5,0.5}
\definecolor{codepurple}{rgb}{0.58,0,0.82}
\definecolor{backcolour}{rgb}{0.95,0.95,0.92}

\lstdefinestyle{mystyle}{
	backgroundcolor=\color{backcolour},   
	commentstyle=\color{codegreen},
	keywordstyle=\color{magenta},
	numberstyle=\tiny\color{codegray},
	stringstyle=\color{codepurple},
	basicstyle=\ttfamily\footnotesize,
	breakatwhitespace=false,         
	breaklines=true,                 
	captionpos=b,                    
	keepspaces=true,                 
	numbers=left,                    
	numbersep=5pt,                  
	showspaces=false,                
	showstringspaces=false,
	showtabs=false,                  
	tabsize=2
}

\usepackage{makecell}

\usepackage[T1]{fontenc}
\usepackage[utf8]{inputenc}
\usepackage{authblk}
\title{
Bridging Performance Gaps for ECG Foundation Models: A Post-Training Strategy
}

\author[1,\#,*]{Ya Zhou}
\author[1,\#]{Yujie Yang}
\author[2,3,*]{Xiaohan Fan}
\author[4,*,\dag]{Wei Zhao}
\affil[1]{Department of Information Center, Fuwai Hospital, Chinese Academy of Medical Sciences and Peking Union Medical College, Beijing, 100037, China}
\affil[2]{Function Test Center, Fuwai Hospital, National Center for Cardiovascular Diseases, Chinese Academy of Medical Sciences and Peking Union Medical
	College, Beijing, 
	China,  100037, China}
\affil[3]{
Cardiac Arrhythmia Center, Fuwai Hospital, National Center for Cardiovascular Diseases, Chinese Academy of Medical Sciences and Peking Union Medical College, Beijing, 
	China,  100037, China}
\affil[4]{Center for Health Statistics and Information, National Health Commission People's Republic of China, Beijing, 100044, China}

\date{}

\begin{document}
\maketitle

\def\thefootnote{\#}\footnotetext{These authors contributed equally.}
\def\thefootnote{*}\footnotetext{Corresponding authors: Wei Zhao (zhaowei@nhc.gov.cn), Xiaohan Fan (fanxiaohan@fuwaihospital.org), Ya Zhou (zhouya@fuwai.com) }\def\thefootnote{\arabic{footnote}}
\def\thefootnote{ \dag}\footnotetext{Supervision: Wei Zhao}

\begin{abstract}
ECG foundation models are increasingly popular due to their adaptability across various tasks. However, their clinical applicability is often limited by performance gaps compared to task-specific models, even after pre-training on large ECG datasets and fine-tuning on target data. This limitation is likely due to the lack of an effective post-training strategy. In this paper, we propose a simple yet effective post-training approach to enhance ECG foundation models. We evaluate it on a publicly available Transformer-based foundation model.  Experiments across multiple ECG tasks show that our method consistently outperforms baseline fine-tuning. On the PTB-XL benchmarks, it improves macro AUROC by 0.7\%-8.9\% and macro AUPRC by 23.3\%-77.9\%, also outperforming several recent state-of-the-art approaches,  including task-specific and advanced architectures.  Further analyses demonstrate improved training dynamics and data efficiency, with only 30\% of the training data outperforming the baseline trained on the full dataset. Ablation studies highlight the importance of stochastic depth and preview linear probing. These findings underscore the potential of post-training strategies to improve ECG foundation models, and we hope this work will contribute to the continued development of foundation models in the ECG domain.
\end{abstract}

\section{Introduction}
\label{sec:intro}
Foundation models have recently emerged as a transformative paradigm in medical artificial intelligence, enabling the development of generalizable and transferable models across diverse clinical tasks \citep{moor2023foundation, bommasani2021opportunities}. Leveraging large-scale pretraining, these models can be efficiently adapted to various downstream applications with limited task-specific data, showing great promise in biomedical analysis. In the electrocardiogram (ECG) domain, several foundation models, such as ECG-FM \citep{mckeen2024ecg}, HuBERT-ECG \citep{coppola2024hubert}, and ST-MEM \citep{na2024guiding}, have been introduced to advance ECG-based tasks. Trained on extensive ECG datasets, these models aim to capture universal cardiac representations and support multiple downstream applications, such as arrhythmia classification \citep{hannun2019cardiologist, ribeiro2020automatic, jiang2024self} and disease risk prediction \citep{somani2021deep, siontis2021artificial, poterucha2025detecting, kwon2020deep, kulkarni2023machine}.

Despite their promise, ECG foundation models still face notable limitations. When fine-tuned for specific diagnostic tasks, their performance often lags behind task-specific or advanced architectures such as multi-scale \citep{shi2023sequence} and state-space models \citep{zhang2023effectively, behrouz2024chimera}. For instance, HuBERT-ECG-BASE performs competitively across several PTB-XL tasks \citep{strodthoff2020deep} yet remains inferior to many specialized models \citep{coppola2024hubert}. Benefiting from the recent open-source releases of many ECG foundation models, we were able to conduct a series of experiments on the PTB-XL dataset. Our results revealed similar findings: although ECG foundation models demonstrate broad adaptability across diverse tasks, their fine-tuning performance remains suboptimal when transferred to datasets from different sources, highlighting concerns about their practical applicability in real-world clinical settings.

To address these limitations, we propose a simple yet effective post-training strategy to enhance ECG foundation models. Using the performance of a Transformer-based model \citep{na2024guiding} on the PTB-XL all-label classification benchmark as an example, we demonstrate that our strategy yields remarkable performance gains of 5.2\% in macro AUROC and 34.9\% in macro AUPRC. 
 Furthermore, compared with recent state-of-the-art task-specific and advanced architectures, our approach demonstrates superior classification performance across multiple tasks.
 These results underscore the complementary advantages of the proposed post-training strategy, demonstrating its potential to address the performance gaps observed in ECG foundation models. 


%
%

The proposed post-training strategy is grounded in two key insights. First, ECG signals inherently exhibit information redundancy, where neighboring time points and heartbeat cycles are often predictable from each other \citep{zhou2025enhancing}. To address this, we employ stochastic depth \citep{huang2016deep} to reduce redundancy and enhance the model’s robustness. Second, pre-training provides a valuable initialization that generally improves model performance. However, while the early layers of the model retain the pre-trained weights, the final classification head is typically initialized randomly. To optimize this, we introduce preview linear probing during post-training to better initialize the classification head, thereby boosting task-specific performance. Although these two components—stochastic depth and preview linear probing—are relatively simple, they play a crucial role in driving improvements in the performance of the ECG foundation model.


In summary, the main contributions of this work are as follows:

\begin{itemize}
	\item [1. ] \textbf{Post-Training Strategy for General ECG foundation models}.  
	We propose a simple yet effective post-training strategy to enhance ECG foundation models. This strategy serves as a crucial complement to existing fine-tuning frameworks, enabling more efficient and robust adaptation across diverse downstream ECG classification tasks.
	
	\item [2. ] \textbf{Significant Performance Gains}.  	
	Experiments demonstrate that our approach consistently outperforms the baseline fine-tuning strategy across multiple datasets. On the PTB-XL benchmarks, our method achieves improvements of 0.7\%-8.9\% in macro AUROC and 23.3\%-77.9\% in macro AUPRC, also outperforming many recent state-of-the-art approaches, including task-specific and advanced architectures. 
	
		\item [3.] \textbf{Practical Implications}.  
	Recent studies have reported sub-optimal performance of fine-tuned ECG foundation models on the PTB-XL benchmark, raising concerns regarding their clinical applicability. 
	Our results suggest that this limitation likely stems from the absence of an effective post-training strategy, which could address the gap and enable real-world clinical adoption.
	
	
	%
	
%
%
%
\end{itemize}

\section{Related Work}
\label{sec:related_work}
\subsection{ECG Classification and Evaluation Benchmarks}
\label{subsec:ecg_class}
\label{subsec:benchmark}
ECG classification is one of the most important tasks in ECG analysis, crucial for many critical applications, including the detection of arrhythmia \citep{hannun2019cardiologist, ribeiro2020automatic, hughes2021performance}, structural heart disease \citep{poterucha2025detecting}, anemia \citep{kwon2020deep}, and diabetes \citep{kulkarni2023machine}. Due to their excellent performance in both accuracy and diagnostic speed \citep{jiang2024self}, deep learning methods have gained significant popularity in these tasks. Successful model architectures in these areas range from convolutional neural networks (CNNs)  and Transformer to state-space models and ensemble approaches \citep{strodthoff2020deep, cheng2023msw, shi2023sequence, zhang2023effectively, behrouz2024chimera}. The task-specific and advanced architectures have proven effective in modeling the time-series nature of ECG signals, including the residual models like resnet1d101 \citep{strodthoff2020deep}, resnet1d\_wang \citep{wang2017time}, the
multi-scale models like MSW-Transformer \citep{cheng2023msw} and MULTIRESNET \citep{shi2023sequence}, as well as advanced state-space models like SPACETIME \citep{zhang2023effectively} and Chimera \citep{behrouz2024chimera}.

Despite rapid methodological progress, fair comparison across ECG classification methods remains challenging due to the lack of benchmark datasets. Although numerous large-scale ECG datasets have been collected worldwide, many are not publicly available \citep{somani2021deep, siontis2021artificial}. To address this limitation, several large open-access ECG datasets have been released in recent years \citep{clifford2024past}, including PTB-XL \citep{wagner2020ptb}, Chapman-Shaoxing \citep{zheng202012}, and Ningbo \citep{zheng2020optimal}. Among these, PTB-XL has become the one of the most widely adopted benchmarks due to its comprehensive usage guidelines, including predefined train-test splits, standardized label sets, and recommended evaluation metrics \citep{strodthoffopen}. These resources have established PTB-XL as a widely acceptable benchmark for ECG classification, enabling reproducible and comparable evaluation across diverse studies \citep{strodthoffopen, strodthoff2020deep}.

\subsection{ECG Foundation Models}
\label{subsec:ecg_foundation}

Recent advances in foundation models for natural language processing have inspired similar approaches for health data \citep{clifford2024past}. In the ECG domain, several studies have explored building foundation models through large-scale pre-training followed by evaluation across diverse downstream tasks. Recent examples include ECG-FM \citep{mckeen2024ecg}, HuBERT-ECG \citep{coppola2024hubert}, and ECGFounder \citep{li2025electrocardiogram}. Noting that the predominant architecture of existing foundation models is Transformer-based, this paper focuses on Transformer-based models, although the proposed strategy may also be applicable to CNN-based models, such as ECGFounder.
Specifically, ECG-FM is a Transformer-based model pre-trained on over 1 million 12-lead ECGs from multiple sources; HuBERT-ECG is another Transformer-based model pre-trained on more than 9 million 12-lead ECGs covering 164 cardiovascular conditions. Another line of ECG foundation models has emerged from research on self-supervised learning for ECG signals. For instance, \cite{na2024guiding} proposed ST-MEM, a Transformer-based SSL framework that explicitly captures the spatiotemporal dependencies inherent in ECG signals. ST-MEM was pre-trained on 188,480 ECGs and evaluated across various downstream tasks. Notably, all four models have publicly released their source code and pre-trained weights, promoting transparency and reproducibility in ECG foundation model research.

Comparing the performance of different ECG foundation models would provide valuable insights for practitioners. However, conducting a fair comparison is challenging, especially for ECG foundation models due to their higher computational cost. As discussed in Section \ref{subsec:benchmark}, the PTB-XL benchmark offers a fair platform for evaluating various algorithms. Nonetheless, the pre-training dataset for ECG-FM uses random splits of the PhysioNet Challenge 2021 dataset \citep{reyna2021will, goldberger2000physiobank}, which includes the PTB-XL dataset, potentially limiting its applicability for evaluation on the PTB-XL benchmark. 
By contrast, HuBERT-ECG reports benchmark results directly on PTB-XL, enabling more straightforward comparisons. In this work, we focus on developing a general post-training strategy for ECG foundation models. Given that Transformer-based architectures are widely adopted in contemporary foundation models, we select ST-MEM \citep{na2024guiding} as a representative example. Throughout this paper, we denote such Transformer-based foundation models as Transformer-FM, emphasizing that the proposed post-training strategy is general and not specifically tailored to any single model.


 

\subsection{Regularization in Deep Learning}
\label{subsec:reg}

Deep learning models often require significant capacity to capture complex representations, making them susceptible to overfitting, especially when the available training data is limited.  Regularization is a widely used technique to mitigate overfitting and improve model generalization. In general, regularization refers to any technique applied in the model architecture, learning process, or inference to alleviate the limitations of insufficient training data \citep{moradi2020survey}. A common regularization method is weight decay \citep{hanson1988comparing}, which penalizes large weights during each update step. \cite{loshchilov2018decoupled} demonstrated that weight decay, combined with a cosine annealing learning rate schedule \citep{loshchilov2017sgdr}, benefits optimizers like Adam \citep{kingma2014adam}. Similarly, \cite{krizhevsky2012imagenet} showed that weight decay, when combined with Dropout \citep{hinton2012improving, srivastava2014dropout}, reduces overfitting effectively.  In addition, stochastic depth is another regularization technique designed to address challenges like vanishing gradients, diminishing feature reuse, and long training times in deep networks \citep{huang2016deep}. Recent work \citep{zhou2025enhancing} demonstrated that stochastic depth is also effective in reducing information redundancy in ECG signals, improving performance on various ECG classification tasks. We explore these strategies in our study.

Following the general definition of regularization, initialization strategies can be considered a form of regularization. The pre-training paradigm, which has been successful in many ECG tasks \citep{na2024guiding, coppola2024hubert, mckeen2024ecg}, retains the pre-trained weights in the early layers of the model. However, the final classification head is typically initialized randomly. In this paper, we use linear probing to initialize this layer. Furthermore, early stopping can also be regarded as a regularization technique, and it is often used alongside other regularization methods \citep{goodfellow2016deep}. We incorporate early stopping in our experiments as well.

\section{Methodology}
\label{sec:method}

In this study, we propose a simple yet effective post-training strategy to enhance the classification performance of ECG foundation models. 
The proposed strategy is model-agnostic, and a Transformer-based foundation model is selected as a representative example to demonstrate its effectiveness. 
Notably, the components of the proposed strategy are easy to implement and can be readily integrated into a wide range of ECG models.

\begin{figure*}[htbp]
	\centering
	\includegraphics[width=0.8\textwidth]{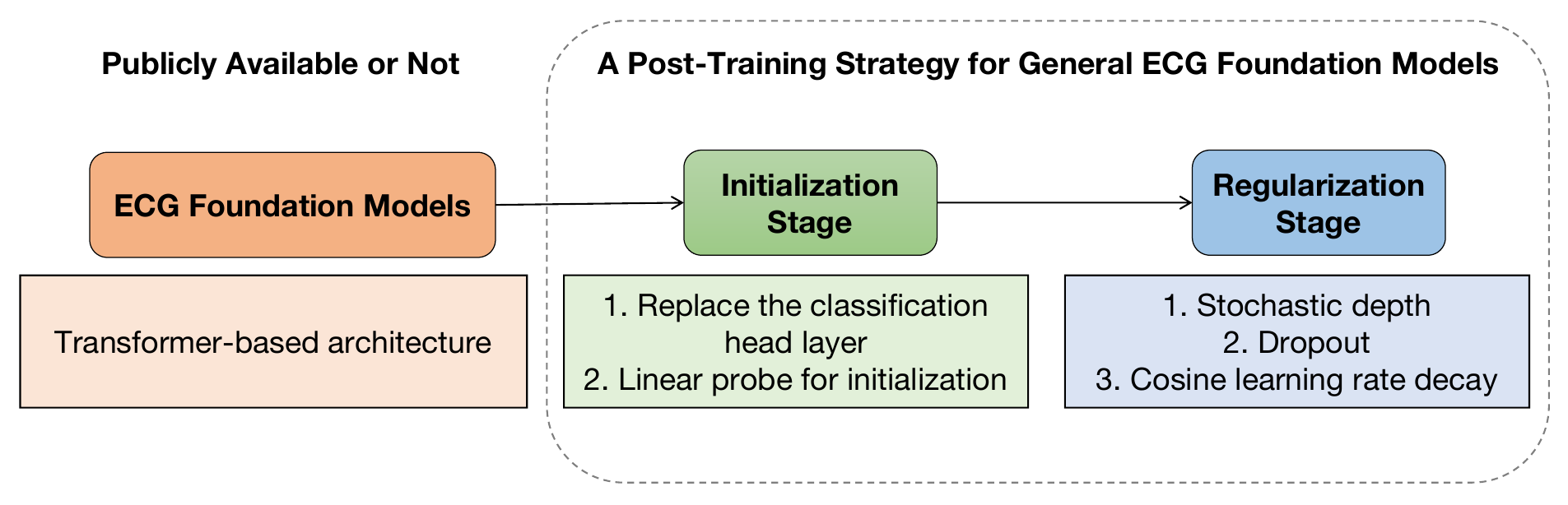}
	\caption{Overview of the proposed post-training framework for ECG classification.}
	\label{fig:framework}
\end{figure*}

 The overall framework of the proposed approach is illustrated in Figure \ref{fig:framework}. It builds upon pre-trained ECG foundation models, and introduces two key stages for downstream classification: an initialization stage and a regularization stage. The design and implementation of each stage are described in detail in the following subsections.


\subsection{Overview of the Transformer-based example}

Transformer-FM \citep{na2024guiding} is a Transformer-based foundation model adapted from the Masked Autoencoder (MAE) framework \citep{he2021masked}, specifically designed to capture the spatio-temporal relationships in multi-lead ECG signals. It segments ECG sequences along both the temporal and lead dimensions, employing a lead-wise shared decoder and lead-specific embeddings to effectively model inter-lead dependencies. For pre-training, Transformer-FM was trained on a hybrid dataset containing 188,480 ECGs from the Chapman-Shaoxing \citep{zheng202012}, Ningbo \citep{zheng2020optimal}, and CODE-15 \citep{ribeiro2021code} datasets. For downstream tasks, it supports both linear probing and full fine-tuning, following the MAE training protocol with a cosine learning rate decay and the AdamW optimizer.

\subsection{Post-training Strategy}
As illustrated in Figure \ref{fig:framework}, our method builds upon pre-trained ECG foundation models and introduces a two-stage post-training strategy. The strategy is designed to enhance classification performance while requiring minimal modifications to the original model architecture. Specifically, it consists of an initialization stage and a regularization stage, which are described in detail below.

\subsubsection{Initialization Stage}  
The initialization stage follows a standard linear probing approach. Specifically, all pretrained backbone parameters are frozen, and the linear classification head is replaced with a task-specific counterpart trained from scratch. This design allows the final layer to be rapidly aligned with the pretrained representations while providing a well-conditioned initialization for downstream classification. As discussed in Section \ref{sec:intro}, an appropriate initialization can facilitate task-specific adaptation and ultimately improve model performance. Following the general definition of regularization presented in Section \ref{sec:related_work}, this step can also be viewed as a form of regularization.

\subsubsection{Regularization Stage}  
The regularization stage involves full fine-tuning of the entire model following initialization, with multiple regularization techniques incorporated to improve generalization. The core component is the stochastic depth strategy \citep{huang2016deep}, which has been shown to effectively reduce redundancy in ECG signals \citep{zhou2025enhancing}. Specifically, stochastic depth is applied to all residual connections in Transformer-FM. 
To further stabilize training, Dropout \citep{hinton2012improving, srivastava2014dropout} is applied to the feed-forward linear layers for Transformer-FM. In addition, a cosine annealing learning rate schedule \citep{loshchilov2017sgdr} and weight decay regularization \citep{hanson1988comparing, loshchilov2018decoupled} are employed to ensure smooth learning rate decay and enhance optimization robustness for Transformer-FM.

\section{Experiments}
\label{sec:exp}

This section presents a comprehensive evaluation of the proposed post-training strategy. We first describe the PTB-XL benchmark dataset \citep{strodthoff2020deep}, which serves as the primary evaluation platform.
We then detail the experimental settings, followed by comparative analyses against the baseline fine-tuning strategies proposed in the original ECG foundation model papers \citep{na2024guiding} and several state-of-the-art methods, including both foundation models and task-specific approaches. 


\subsection{Datasets}
\label{subsec:dataset}

The PTB-XL database is a widely used open-source dataset for evaluating ECG model performance, notable for its relatively large sample size, high-quality annotations and clearly defined usage guidelines \citep{wagner2020ptb, strodthoffopen}. It contains 21,837 clinical 12-lead ECG recordings, each 10 seconds in duration, collected from 18,885 patients across a broad age range (0-95 years). Each recording was annotated by up to two cardiologists, with labels drawn from a set of 71 distinct ECG statements. These statements can be organized into 12 rhythm labels and 44 diagnostic labels, which can further be hierarchically grouped into 24 subclasses. Multiple labels may be associated with a single recording, reflecting the inherently multi-label nature of clinical ECG interpretation. Based on this labeling system, we consider four multi-label classification tasks: (1) all-71, which uses the complete set of 71 labels; (2) rhythm-12, which focuses on 12 rhythm categories; (3) diagnostic-44, which focuses on 44 diagnostic categories; and (4) subclass-23, which focuses on 23 hierarchical subclasses derived from the diagnostic labels. As discussed in Section \ref{sec:related_work}, PTB-XL provides predefined splits into training, validation, and testing sets (approximately 8:1:1), enabling consistent and fair comparisons across methods \citep{strodthoff2020deep, strodthoffopen}. We follow this standard data splitting  in this study.

\subsection{Experimental Settings}
\label{subsec:exp_setup}

%

In the initialization stage, we adopt the same linear probing configuration as in \citep{na2024guiding}, training for 100 epochs, since the validation AUROC convergence for Transformer-based models generally requires a longer schedule. In the regularization stage, we retain the fine-tuning hyperparameters from \citep{na2024guiding}, but incorporate a stochastic depth rate of 0.1 and a dropout rate of 0.01 to enhance robustness and prevent overfitting. The model is evaluated on the validation set after each epoch, and the checkpoint with the highest validation macro AUROC is used for final testing. This strategy serves as a form of early stopping regularization.

\subsection{Comparison with Other Methods}
\label{subsec:com}
This section presents a comprehensive comparison between the proposed post-training method and a range of baseline approaches, including the suggested baseline fine-tuning strategy, referred to Transformer-FM \citep{na2024guiding}. To distinguish the post-training variant, we denote our method as Transformer-FM-PT. Additional baselines include the foundation model HuBERT-ECG-BASE \citep{coppola2024hubert} and task-specific and advanced models, such as inception1d \citep{ismail2020inceptiontime}, resnet1d101 \citep{strodthoff2020deep}, resnet1d\_wang \citep{wang2017time}, ensemble-7-models \citep{strodthoff2020deep}, SPACETIME \citep{zhang2023effectively}, MSW-Transformer \citep{cheng2023msw}, MULTIRESNET \citep{shi2023sequence}, and Chimera \citep{behrouz2024chimera}. The reported performance values are derived from \cite{strodthoff2020deep, zhang2023effectively, cheng2023msw, shi2023sequence, coppola2024hubert, behrouz2024chimera}.

\begin{table*}[!htbp]
	\small
	\centering
		\caption{Performance comparison in terms of macro AUROC. The first column indicates whether a method is an ECG foundation model, and the third column specifies the main model architecture. Best results are highlighted in bold.}
			\scalebox{0.7}{
	\begin{tabular}{c lccccc}
		\toprule
		\textbf{Foundation Model} &	\textbf{Method} & \textbf{Architecture} & \textbf{all-71} & \textbf{diagnostic-44} & \textbf{subclass-23} & \textbf{rhythm-12} \\
		\midrule
		No &	inception1d & CNN &0.925 & 0.931 & 0.930 & 0.953 \\
		No &	xresnet1d101 & CNN&0.925 & 0.937 & 0.929 & 0.957 \\
		No &	resnet1d\_wang &CNN &0.919 & 0.936 & 0.928 & 0.946 \\
		No &	ensemble-7-models & Ensemble (CNN + others) &0.929 & 0.939 & 0.933 & 0.965 \\
		No &	MULTIRESNET & CNN (multi-scale) &0.938 & 0.939 & 0.934 & \textbf{0.975}\\
		No &	MSW-Transformer & Transformer (multi-scale)& - & 0.924 & 0.927 & 0.925  \\
		No &	SPACETIME & State space model & 0.936 & 0.941 & 0.933 & 0.967 \\
		No &	Chimera & State space model & {0.941} & \textbf{0.947} & {0.935} & \textbf{0.975} \\

		Yes &	HuBERT-ECG-BASE & Transformer & 0.903 & 0.917 & 0.917 & 0.953\\
		\midrule
		
	Yes & Transformer-FM
		& Transformer & \makecell{0.898 \\ (0.888, 0.907)} 
		& \makecell{0.870 \\ (0.856, 0.884)} 
		& \makecell{0.910 \\ (0.896, 0.922)} 
		& \makecell{0.968 \\ (0.942, 0.984)} \\
	Yes & Transformer-FM-PT
		& Transformer & \makecell{ \textbf{0.945} \\ (0.938, 0.951)} 
		& \makecell{\textbf{0.947} \\ (0.938, 0.954)} 
		& \makecell{\textbf{0.943} \\ (0.933, 0.952)} 
		& \makecell{\textbf{0.975} \\ (0.956, 0.987)} \\
		\bottomrule
	\end{tabular}}

	\label{tab:performance}
\end{table*}

%

Table \ref{tab:performance} presents the macro AUROC of all methods, along with 95\% confidence intervals computed using 1000 bootstrapped samples. Transformer-FM-PT achieves the highest AUROC on four tasks: all-71 (0.945), diagnostic-44 (0.947) and subclass-23 (0.943), and  rhythm-12 (0.975). 
Compared with the suggested fine-tuning strategy Transformer-FM, the proposed post-training strategy improves performance by 0.7\%-8.9\%, respectively, demonstrating the effectiveness of the post-training adaptation.

Beyond outperforming the baseline fine-tuning methods, the foundation models enhanced with our post-training strategy also surpass several recent state-of-the-art task-specific approaches. Compared to other CNN-based baselines (inception1d, resnet1d101, resnet1d\_wang, and ensemble-7-models) from \cite{strodthoff2020deep}, Transformer-FM-PT yields macro AUROC gains of 0.9\%-3.1\%. When compared to multi-scale architectures like MULTIRESNET and MSW-Transformer, which explicitly exploit the hierarchical properties of ECG signals, it improves performance by 0.7\%-2.5\%, except for rhythm-12, where MULTIRESNET achieves the same performance. Even compared with recent state space models (SPACETIME and Chimera), Transformer-FM-PT achieves consistent gains of 0.4\%-1.1\%, except for diagnostic-44 and rhythm-12, where Chimera achieves the same performance. Collectively, these results highlight the robustness of post-training strategy and its ability to bridge the adaptation gap of foundation models, achieving balanced generalization across diverse ECG classification tasks.

Although AUROC is a standard benchmark metric, it may be less sensitive in the presence of class imbalance. To address this, we additionally report macro Area Under the Precision-Recall Curve (AUPRC) in Table \ref{tab:performance2}, with 95\% confidence intervals computed using 1000 bootstrapped samples. Since AUPRC results for other methods are not available, the comparison is limited to Transformer-FM. Across all tasks, the proposed post-training strategy consistently surpasses the baseline, achieving relative improvements of 23.3\%-77.9\%, underscoring the effectiveness of the post-training strategy, particularly in imbalanced data scenarios. 




\begin{table*}[!htbp]
	\small
		\caption{Performance comparison in terms of macro AUPRC. Best  results are highlighted in bold.}
	\centering
		\scalebox{0.7}{
	\begin{tabular}{lcccc}
		\toprule
		\textbf{Method} & \textbf{all-71} & \textbf{diagnostic-44} & \textbf{subclass-23} & \textbf{rhythm-12} \\
		\midrule
		Transformer-FM
		& {0.307  (0.305, 0.337)} 
		& {0.231 (0.224, 0.266)} 
		& {0.441 (0.419, 0.488)} 
		& {0.559 (0.520, 0.656)} \\
		Transformer-FM-PT
		& {\textbf{0.414}  (0.405, 0.452)} 
		& {\textbf{0.411} (0.393, 0.453)} 
		& {\textbf{0.555} (0.531, 0.592)} 
		& {\textbf{0.689} (0.623, 0.774)} \\
		\bottomrule
	\end{tabular}
}
	\label{tab:performance2}
\end{table*}

\subsection{Per-Label Performance Evaluation}
\begin{figure*}[!htp]
	\centering
	\subfloat[ ]{%
		\includegraphics[width=0.95\textwidth]{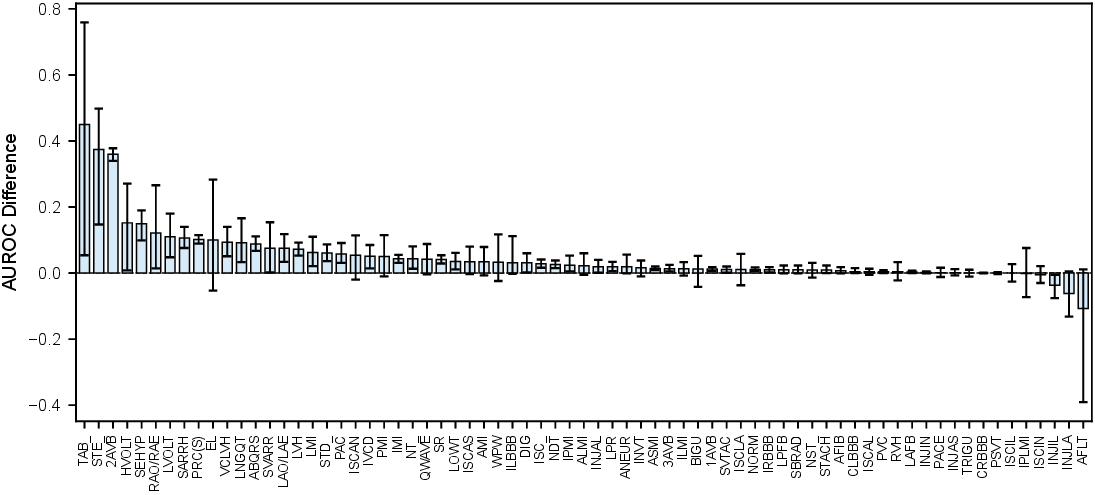}%
		\label{fig:auroc_difference}%
	}\\[0.2cm]  
	\subfloat[ ]{%
		\includegraphics[width=0.95\textwidth]{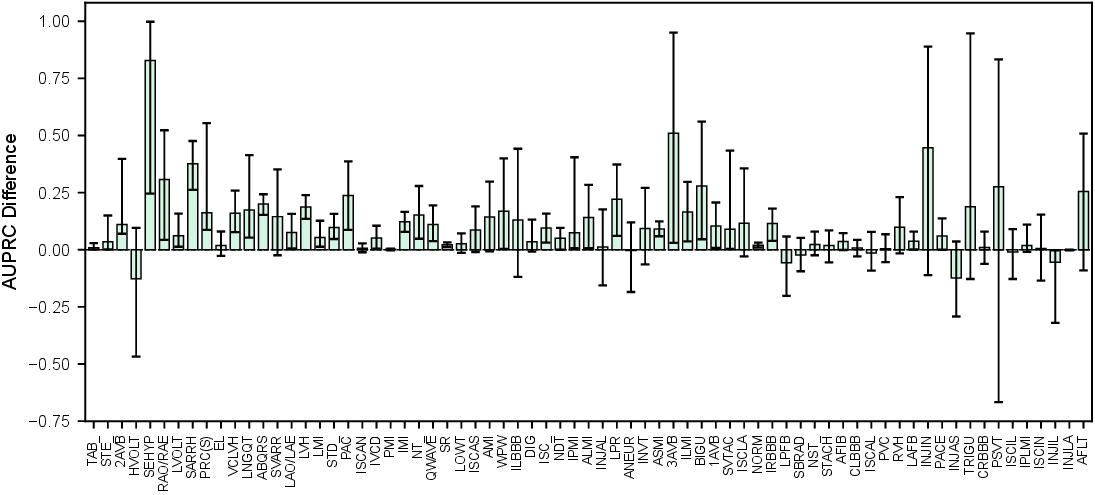}%
		\label{fig:auprc_difference}%
	}
\caption{ Per-label comparison of AUROC and AUPRC differences between the proposed post-training strategy and the baseline (Transformer-FM-PT $-$ Transformer-PT) across 71 ECG diagnoses. Bars represent the observed differences on the test set, and vertical error bars indicate the 95\% confidence intervals estimated from 1,000 bootstrap resamples.}
	\label{fig:per_label_performance}
\end{figure*}

To gain a more detailed understanding of the effectiveness of our proposed post-training strategy, we evaluated both AUROC and AUPRC for each of the 71 ECG diagnoses on the test set. Figures \ref{fig:auroc_difference} and \ref{fig:auprc_difference} present the per-label AUROC and AUPRC differences between our post-training method and the baseline (Transformer-FM-PT $-$ Transformer-PT). In these figures, bars represent the observed differences on the test set, while vertical error bars indicate the 95\% confidence intervals obtained from 1,000 bootstrap resamples. For AUROC, 63 out of 71 labels exhibited positive differences, with 35 labels showing statistically significant improvements (i.e., the lower bound of the 95\% confidence interval is greater than zero). Similarly, for AUPRC, 62 labels demonstrated positive differences, with 35 reaching statistical significance. For the remaining labels, most confidence intervals spanned both positive and negative values, indicating that the observed differences were not statistically significant. 

Overall, these results demonstrate that our post-training strategy consistently enhances predictive performance across the majority of ECG diagnoses, providing a more granular than the macro-level metrics reported in the previous section.

\subsection{Convergence Analysis}

\begin{figure*}[!htp]
	\centering
	\subfloat[ ]{%
		\includegraphics[width=0.48\textwidth]{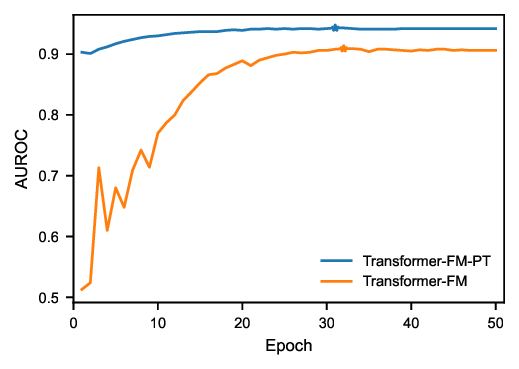}%
		\label{fig:convergence_auroc}%
	}
	\hfill
	\subfloat[ ]{%
		\includegraphics[width=0.48\textwidth]{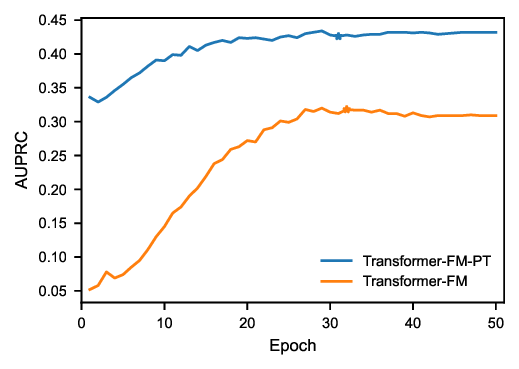}%
		\label{fig:convergence_auprc}%
	}
	\caption{
Validation set performance of the proposed post-training strategy compared with the baseline over 50 training epochs for the all-71 task. (a) AUROC and (b) AUPRC are shown for both methods.
The early stopping epoch, determined based on the validation AUROC, is highlighted with a star marker.
	}
	\label{fig:convergence_analysis}
\end{figure*}

To evaluate the convergence behavior of the proposed post-training strategy (Transformer-FM-PT) compared with the baseline (Transformer-FM), we conducted experiments on the validation set over 50 training epochs for the all-71 task. As shown in Figures \ref{fig:convergence_auroc} and \ref{fig:convergence_auprc}, both methods exhibit similar training dynamics: performance increases rapidly during the early epochs, reaches a plateau in the middle epochs, and remains relatively stable toward the end of training. Specifically, the proposed method surpasses the peak performance of the baseline within the first 4 epochs, achieving AUROC = 0.912 and AUPRC = 0.346, whereas the baseline reaches its maximum AUROC = 0.909 and AUPRC = 0.318 only later. From epoch 21 onward, the proposed method maintains AUROC values above 0.940 and AUPRC values above 0.420, indicating stable predictive accuracy in the later stages of training. Collectively, these results suggest that the proposed post-training strategy facilitates rapid attainment of strong validation performance while preserving stable optimization behavior.

\subsection{Data Efficiency}
\label{subsect:varying_sample}

\begin{figure*}[!bhtp]
	\centering
	\subfloat[ ]{
		\includegraphics[width=0.48\textwidth]{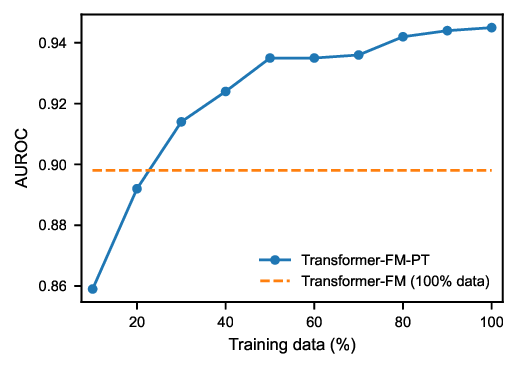}%
		\label{fig:learning_curve_auroc}%
	}
	\hfill
	\subfloat[ ]{
		\includegraphics[width=0.48\textwidth]{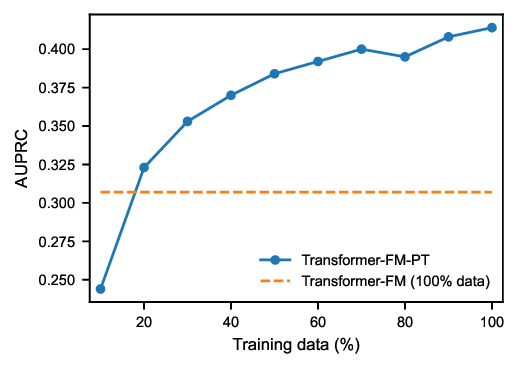}%
		\label{fig:learning_curve_auprc}%
	}
	\caption{
	Test set performance of the proposed post-training strategy under varying proportions of training data.  
	(a) AUROC and (b) AUPRC are reported for 10\%, 20\%, ..., up to 100\% of the training data.  The dashed horizontal line indicates the baseline performance trained using 100\% of the training data.  
	}
	\label{fig:learning_curves}
\end{figure*}

To evaluate the robustness and data efficiency of the proposed post-training strategy under varying amounts of training data, we trained the model using 10\%, 20\%, \ldots, up to 100\% of the available training samples for the all-71 task, and evaluated performance on the test set in terms of AUROC and AUPRC. As shown in Figures \ref{fig:learning_curve_auroc} and \ref{fig:learning_curve_auprc}, both metrics generally improve as the amount of training data increases and reach their highest values when the full training set is used. Notably, the proposed method achieves superior performance even with limited training data, outperforming the baseline trained on the full dataset. For example, when trained with only 30\% of the available data, the proposed method attains an AUROC of 0.924 and an AUPRC of 0.370, both exceeding the baseline performance obtained using 100\% of the training data (AUROC = 0.898, AUPRC = 0.307). These results indicate that the proposed post-training strategy is more data-efficient and yields improved generalization in low-data regimes, while consistently maintaining performance advantages across a wide range of training data proportions.

\subsection{Ablation Study}
\begin{figure*}[!hbtp]
	\centering
	\subfloat[ ]{%
		\includegraphics[width=0.5\textwidth]{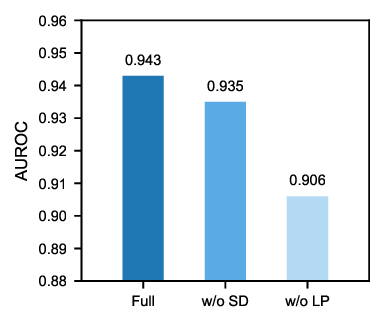}%
		\label{fig:ablation_auroc}%
	}
	\hfill
	\subfloat[ ]{%
		\includegraphics[width=0.5\textwidth]{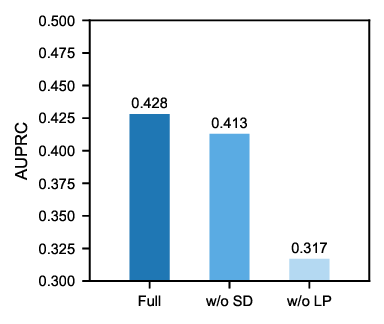}%
		\label{fig:ablation_auprc}%
	}
	
\caption{
	Validation set performance of the proposed post-training strategy for the ablation study on the all-71 task.
	(a) AUROC and (b) AUPRC are shown for different ablation settings. Bars represent performance for each setting, with color intensity reflecting the relative magnitude of each metric. 
	\texttt{Full} denotes using the complete post-training strategy; 
	\texttt{w/o SD} denotes removing the stochastic depth strategy; \texttt{w/o LP} denotes removing the linear probing initialization strategy. 
}
	\label{fig:ablation_study}
\end{figure*}

To evaluate the contribution of the key components in the proposed post-training strategy, namely the preview linear probing (LP) initialization and stochastic depth (SD), we conducted an ablation study on the PTB-XL all-71 task. The results are summarized in Figure \ref{fig:ablation_study}. For AUROC, removing the SD and LP components led to decreases of 0.8\% and 3.9\% on the validation set, respectively (Figure \ref{fig:ablation_auroc}). For AUPRC, the corresponding decreases were 3.5\% and 25.9\% (Figure \ref{fig:ablation_auprc}). These findings demonstrate that, despite the simplicity of the approach, both LP initialization and SD are crucial for enhancing post-training performance.

\section{Discussion}
\label{sec:discussion}
\subsubsection*{Performance Gains on ECG Foundation Models}
In this study, we evaluated the effectiveness of the proposed post-training strategy on four PTB-XL multi-label ECG classification benchmark tasks. The method consistently improves performance over recent ECG foundation models, demonstrating its ability to enhance generalization across diverse ECG classification tasks. As an example, we applied the strategy to a Transformer-based ECG foundation model. The macro AUROC increases by 0.7\%-8.9\% and the macro AUPRC by 23.3\%-77.9\% compared to the original fine-tuning strategy. Per-label analysis revealed performance gains in the majority of ECG diagnoses, with 35 diagnoses exhibiting statistically significant improvements. These results highlight the effectiveness of the proposed post-training approach in enhancing the performance of ECG foundation models.

\subsubsection*{Comparison with State-of-the-Art}
It is worth noting that the proposed method achieves competitive or superior performance compared with several recent state-of-the-art models across diverse ECG classification tasks. Baseline methods such as inception1d, xresnet1d101, resnet1d\_wang, and ensemble-7-models \citep{strodthoff2020deep} exhibit competitive or even superior performance to recent ECG foundation models, likely due to their specialized design for ECG signals. Similarly, the multi-scale architecture MULTIRESNET \citep{shi2023sequence} performs well, reflecting the inherent multi-scale structure in ECG data, while advanced models like SPACETIME \citep{zhang2023effectively} and Chimera \citep{behrouz2024chimera} highlight the significance of carefully designed architectures for ECG time-series data. These task-specific and advanced models reveal a deeper understanding of ECG signals, as they are tailored for the specific challenges of ECG classification. In contrast, foundation models prioritize broad adaptability, which may not rely on a strong inductive bias specific to some ECG tasks. This flexibility allows foundation models to be applied across a wide range of tasks, but in certain cases, it can result in lower performance compared to specialized models, especially in the absence of an appropriate post-training strategy. As demonstrated in both our study and the work of \cite{coppola2024hubert}, foundation models might lag behind task-specific and advanced architectures, even when pre-trained on large ECG datasets. This performance gap underscores the critical need for a robust post-training strategy. Our results, shown in Table \ref{tab:performance}, clearly illustrate that incorporating an effective post-training approach significantly enhances the performance of ECG foundation models, enabling them to match or even surpass task-specific and advanced architectures.

\subsubsection*{Training Dynamics and Data Efficiency} The convergence and data efficiency analyses provide complementary insights into the optimization behavior and practical applicability of the proposed post-training strategy. The example ECG foundation model used in this study contains more than 90M trainable parameters, substantially exceeding the scale of commonly used task-specific ECG classification models~\citep{strodthoff2020deep}. While larger models offer greater representational capacity and adaptability, they also introduce increased challenges in training stability and generalization, particularly in limited-data scenarios. The proposed post-training strategy effectively addresses these challenges, enabling stable optimization and strong performance with reduced training data. Notably, the proposed method surpasses the baseline’s peak validation performance by epoch 4, and when trained with only 30\% of the available data, it outperforms the baseline trained on the full dataset. These results indicate that the proposed strategy improves training dynamics and data efficiency, which is advantageous for real-world clinical applications with constrained annotation budgets.

\subsubsection*{Ablation Study Insights}
This work does not aim to introduce a complex model architecture, but rather focuses on developing effective and principled training strategies. The ablation study demonstrates that removing the two proposed components—linear probing initialization and stochastic depth—leads to notable performance degradation, with decreases of 3.9\% and 0.8\% in macro AUROC, and 25.9\% and 3.5\% in macro AUPRC, respectively. These results indicate that even simple strategies can substantially enhance the performance of deep learning models when properly applied. Indeed, many fundamental components underpinning modern deep learning success, such as residual connections \citep{he2016deep}, batch normalization \citep{ioffe2015batch}, and layer normalization \citep{ba2016layer}, are conceptually simple yet highly effective. Motivated by recent studies in ECG classification \citep{zhou2025enhancing} and sequential downsizing initialization for tensor regression \citep{zhou2024broadcasted}, which emphasize initialization closer to target data distributions, our results suggest that domain-specific observations and insights from traditional statistical models may inspire more principled strategies for improving contemporary deep learning methods.

\subsubsection*{A General Post-training Framework}
It is worth noting that the proposed post-training strategy serves as a general framework applicable to a wide range of ECG foundation models. The proposed initialization strategy for the linear head can be readily applied to almost all ECG foundation models. Similarly, the stochastic depth strategy is broadly applicable, as residual connections are fundamental components in modern neural network architectures. In this study, we use Transformer-FM as an example, primarily because Transformer are most commonly used foundation model architectures. Nevertheless, this does not imply that the proposed post-training strategy is limited to these two models; rather, it provides a generalizable framework that can be easily extended to other ECG foundation models. We encourage the research community to further explore and validate the proposed framework across different architectures and datasets.

\subsubsection*{Evaluation Metrics}
Another important observation pertains to the performance metrics used. While macro AUROC remains the standard metric in ECG benchmarks \citep{strodthoff2020deep}, it is relatively insensitive to class imbalance, which is a common characteristic of clinical ECG datasets. In contrast, macro AUPRC is more sensitive to this imbalance, as it captures the precision-recall trade-off \citep{ozenne2015precision}. This distinction is evident in most of our experiments. For instance, in the four PTB-XL multi-label tasks examined in this study, the proposed method achieves improvements of 0.7\%-8.9\% in macro AUROC compared to the baseline, whereas the improvement in macro AUPRC ranges from 23.3\% to 77.9\%. On the one hand, the substantial gains in macro AUPRC highlight the efficacy of the post-training strategy. On the other hand, the notable discrepancy between macro AUROC and AUPRC underscores an important insight: even a small difference in macro AUROC can correspond to significant variations in macro AUPRC, indicating that different methods may exhibit considerable differences in their ability to detect true positives in the presence of a large number of negatives.

\subsubsection*{Implications and Future Directions}
Our findings have important implications for the design and clinical deployment of ECG foundation models. 
While recent successes of foundation models in natural language processing have inspired similar approaches for health data \citep{clifford2024past}, ECG foundation models face unique challenges due to the domain-specific characteristics of ECG signals, such as high information redundancy and physiological variability \citep{zhou2025enhancing}. 
Recent studies have reported that existing ECG foundation models achieve sub-optimal performance on the PTB-XL benchmark \citep{coppola2024hubert}, raising concerns regarding their direct clinical applicability. 
Our results indicate that this limitation may be attributed to the absence of an effective post-training strategy.
By addressing this gap with our proposed post-training strategy, the ECG foundation model can achieve strong performance on downstream tasks. 
This highlights the critical role of post-training in enabling real-world clinical adoption of foundation models.

\section{Conclusion}
In this study, we propose a simple yet effective post-training strategy for general ECG foundation models. Using a Transformer-based foundation model as an example, we demonstrate the effectiveness of the proposed approach.
Experiments on the PTB-XL dataset show that the proposed method not only significantly outperforms the baseline  fine-tuning but also maintains stable convergence and strong data efficiency.
Moreover, the proposed method achieves competitive or superior performance compared with several recent state-of-the-art models across diverse ECG classification tasks.
These findings underscore the potential of post-training strategies to bridge the adaptation gap for ECG foundation models, highlighting a promising direction to further advance their development.

\bibliographystyle{rss} 
\bibliography{reference}

\end{document}